\documentclass[lettersize,journal,dvipsnames]{IEEEtran}
\usepackage{amssymb,amsmath,bm,bbm,mathtools}
\usepackage{authblk}
\usepackage{amsthm}
\theoremstyle{definition}
\usepackage{amsmath,amsfonts}
\usepackage{algorithmic}
\usepackage{algorithm}
\usepackage{array}
\usepackage{threeparttable}
\usepackage{todonotes}
\usepackage{booktabs}
\allowdisplaybreaks[4]
\usepackage{textcomp}
\usepackage{stfloats}
\usepackage{hyperref,url}
\usepackage{verbatim}
\usepackage{graphicx}
\usepackage{cite}
\usepackage{makecell}
\newtheorem{remark}{Remark}
\usepackage{comment}

\begin{document}

\title{A Physics-Informed Machine Learning for Electricity Markets: A NYISO Case Study}
\author{Robert Ferrando,
Laurent Pagnier,
Robert Mieth,
Zhirui Liang, Yury Dvorkin,
Daniel Bienstock and 
Michael Chertkov}
\markboth{IEEE Transactions on Energy Markets, Policy and Regulation, Vol.~X, No.~Y, March 2023}%
{Shell \MakeLowercase{\textit{et al.}}: A Sample Article Using IEEEtran.cls for IEEE Journals}


\maketitle

\begin{abstract}
This paper addresses the challenge of efficiently solving the optimal power flow problem in real-time electricity markets. The proposed solution, named Physics-Informed Market-Aware Active Set learning OPF (PIMA-AS-OPF), leverages physical constraints and market properties to ensure physical and economic feasibility of market-clearing outcomes. Specifically, PIMA-AS-OPF employs the active set learning technique and expands its capabilities to account for curtailment in load or renewable power generation, which is a common challenge in real-world power systems. The core of PIMA-AS-OPF is a fully-connected neural network that takes the net load and  the system topology as input. The  outputs of this neural network include active constraints such as saturated generators and transmission lines, as well as non-zero load shedding and wind curtailments. These outputs allow for reducing the original market-clearing optimization to a system of linear equations, which can be solved efficiently and yield both the dispatch decisions and the locational marginal prices (LMPs). The dispatch decisions and LMPs are then tested for their feasibility with respect to the requirements for efficient market- clearing results. The accuracy and scalability of the proposed method is tested on a realistic 1814-bus NYISO system with current and future renewable energy penetration levels. 
\end{abstract}


\section{Introduction}
US electricity markets are cleared in a two-stage procedure, including the day-ahead market (DAM) and the real-time market (RTM). The DAM clearing is based on security-constrained unit commitment (SCUC), which determines the least-cost generator schedule for the next day. SCUC considers day-ahead generation offers, load bids, as well as the forecast of load and renewable energy source (RES) generation \cite{DA_scheduling_manual}. The commitment of generators, with the notable exception of flexible, gas-fired units, is predominantly determined in the DAM, but the final RTM production schedule may diverge from the anticipated DAM production schedule due to changes in operating conditions. The cornerstone of both the DAM and RTM clearing is the fast solution of a DC optimal power flow (DC-OPF) problem \cite{RT_scheduling_manual} or one of its variants \cite{li2007dcopf}. Naturally, RTM clears more frequently than DAM and requires the DC-OPF problem to be solved multiple times for each 5- to 15-minute operating interval. Additionally, the drastic growth of RES, especially the recent roll-out of large-scale offshore wind power plants \cite{offshore_wind_report}, has increased the uncertainty in power system scheduling, motivating system and market operators to perform scenario-based uncertainty analyses. These analyses substantially increase the number of DC-OPF-based instances that must be solved within each 5- or 10-minute intervals
This motivates modifying the current ISO procedures to solve DC-OPF more efficiently.

This paper proposes a learning-based algorithm named \textbf{P}hysics-\textbf{I}nformed \textbf{M}arket-\textbf{A}ware \textbf{A}ctive \textbf{S}et learning OPF, or \textbf{PIMA-AS-OPF}. The PIMA-AS-OPF leverages the physical constraints in the DC-OPF model and produces market-clearing outcomes that comply with  market design principles, e.g. revenue adequacy and cost recovery, to ensure both physical and economic feasibility of its outcomes. The accuracy and scalability of the proposed method is tested in a realistic 1814-bus New York Independent System Operator (NYISO) transmission network model, which accounts for the current RES and future higher RES penetration levels.

\subsection{Literature Review} \label{sec:prior}
Machine learning (ML) has been leveraged to expedite the solution of many power systems optimization problems \cite{rahman2020machine}, including DC-OPF \cite{zhang2021convex,chen2022learning,ng2018statistical,ourfirstpaper}, AC-OPF \cite{rahman2020machine,canyasse2017supervised,falconer2022leveraging,lei2020datadriven,nellikkath2022physics}, and security-constrained economic dispatch (SCED) \cite{yang2020fast,Pascal2021}. The ML techniques behind these approaches include logistic regression \cite{canyasse2017supervised},
random forest \cite{rahman2020machine,canyasse2017supervised}, auto-encoders \cite{yang2020fast}, and stacked extreme learning machine \cite{lei2020datadriven}. The most popular technique is the neural network (NN) framework. The NN models designed for solving the power system scheduling models include fully-connected neural networks (FCNN) \cite{chen2022learning,Pascal2021}, input convex neural networks (ICNN) \cite{zhang2021convex}, physics-informed neural networks (PINN) \cite{nellikkath2021physics,nellikkath2022physics}, convolutional neural networks (CNN) \cite{falconer2022leveraging}, graph neural networks (GNN) \cite{falconer2022leveraging,liu2021graph}, and deep neural networks (DNN) \cite{yang2020fast,singh2020learning}.

There are two common ways to construct ML approaches for solving optimization problems. The intuitive way is to train a ML model to predict the solution of the original problem directly based on the inputs. For example, \cite{liu2021graph} proposes a GNN model which takes electricity demands as the input and gives real-time electricity prices as the output. An alternative way is using ML to solve a sub-problem of the original optimization problem, thereby reducing the computational difficulty of the original problem. A common sub-problem is \emph{active set learning}, in which the active inequality constraints of an optimization problem are learned via classification. Reducing the number of constraints under consideration is a key step towards dimension reduction of the optimization problem, and motivates the classification of active constraints, so that optimization theory can be leveraged to alleviate the computational complexity of the problem at hand.

The active set learning technique has been widely used in solving the power system scheduling problem \cite{DeepMisra,misra2022learning,Pascal2021,ourfirstpaper,chen2022learning}. For example, \cite{DeepMisra} proposes using classification algorithms, specifically neural net classifiers, to learn the mapping between uncertainty in electricity demands and the active set of constraints in the DC-OPF model, which enhances computational efficiency. Similarly, \cite{chen2022learning} constructs a NN that predicts the optimal generation cost from the load at each bus, subsequently differentiating the cost with respect to load. The resulting values dual variables (typically interpreted as LMPs) are used to learn the active constraints, from which a system of linear equations equivalent to the original DC-OPF problem can be derived. The current work in \cite{DeepMisra,ourfirstpaper,chen2022learning} sets a precedent for the development and implementation of the active set learning technique on larger systems and more complicated problems, such as AC-OPF and SCED. Inspired by the observation that most of the generators are dispatched to their upper or lower limits, \cite{Pascal2021} develops a ``classification-then-regression'' scheme to solve SCED, i.e., first identifying the generators saturated at their bounds, and then using regression to complete the primal solution with the dispatch of the ``free'' generators . The proposed architecture achieves relative errors below 1\% in numerical experiments on a synthetic French transmission system with realistic Midwest ISO data, demonstrating its potential for real-time energy markets.
Moreover, \cite{yang2020fast} treats SCED, using stacked de-noising autoencoders (SDAEs) to learn the active constraints from ``system operating conditions,'' adapting to changes in grid topology using transfer learning.
As reported in \cite{GOcomp}, one of the winning teams of the ARPA-E GO competition also employed active set learning in their proposed method.
However, as the number of features (buses, lines, etc.) of the power grid increases, so does the number of active constraints and the challenge of the underlying ML problem. To address this, \cite{misra2022learning} proposes an algorithm to learn active constraints irrespective of the optimization problem formulation or its inherent uncertainty. The algorithm in \cite{misra2022learning} offers ``theoretical performance guarantees'' and can quickly converge when there are few active sets to be learned. 
However, most relevant papers, e.g., \cite{canyasse2017supervised,DeepMisra,liu2021graph,chen2022learning}, only evaluate the accuracy of the forecast results (such as the dispatch or prices in electricity markets) using the difference between the forecast and true values, while the feasibility of the forecast results in the actual market setting is ignored. For example, if the power suppliers cannot recover their power generation costs with the forecast real-time energy prices, then the prices are infeasible even if the forecast errors are small in general. On the other hand, most of the DC-OPF models studied in previous work, e.g., \cite{zhang2021convex,chen2022learning,ng2018statistical,ourfirstpaper}, ignore the possible curtailment in load or RES power, which is a common issue in real-world power systems.
In addition, as pointed out in \cite{Pascal2021}, most published case studies are for ``small academic test systems,'' such as IEEE 118-bus system \cite{zhang2021convex}, IEEE 300-bus system \cite{pan2020deepopf}, 500-bus transmission network \cite{rahman2020machine}, or 661-bus utility system \cite{yang2020fast}. These systems are too small to assess scalability of the algorithms therein to larger operational transmission grids.
The work in \cite{Pascal2021} does use a realistically sized test system, but with spatially inconsistent data sets to model RES.

\begin{figure}[h]
\centering
\includegraphics[width=\columnwidth]{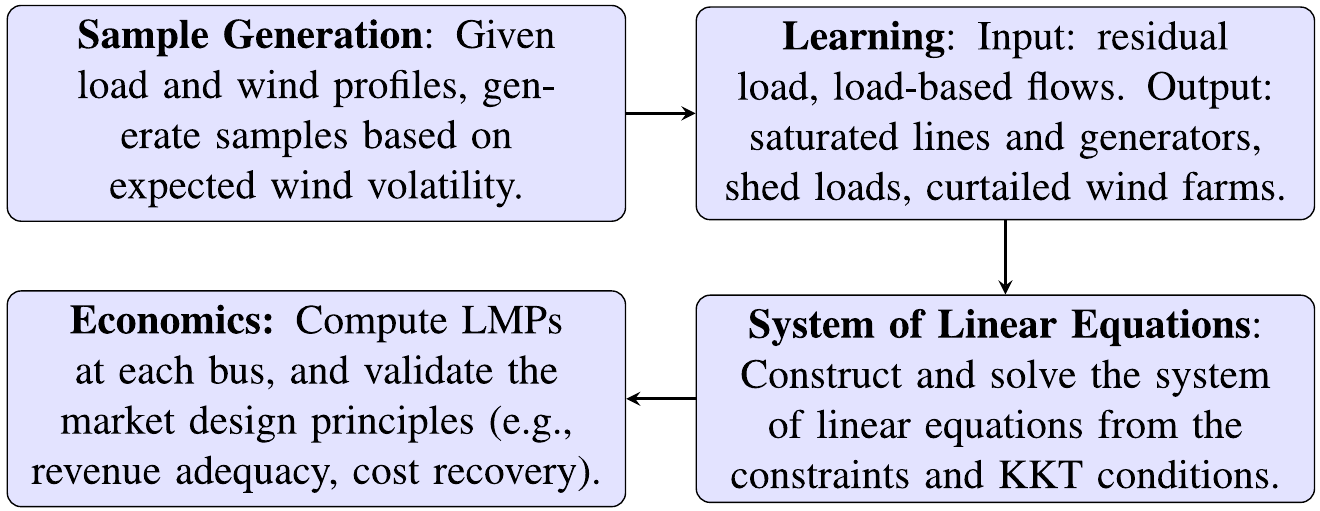}
\caption{A summary of the PIMA-AS-OPF procedure.}
\label{fig:schematic}
\end{figure}

\subsection{Outline of the paper}
This paper proposes a ML-based optimization algorithm, PIMA-AS-OPF, to expedite the solution of DC-OPF in a principled manner. A schematic for the proposed PIMA-AS-OPF approach is shown in Fig.~\ref{fig:schematic}.
First, the day-ahead security constrained unit commitment (SCUC) problem is solved. The real-time DC-OPF problem is then solved to generate samples, with the SCUC solution assumed throughout. For each hour of the next day, the second stage trains specialized ML models (fully-connected neural networks).
The hourly resolution matches market operating practices and aids the reduction of computational complexity, as all real-time problems solved during an hour rely on the same SCUC solution \cite{Pascal2021}.
The NN input is the net load (i.e., demand minus RES injection) and the power flow across transmission lines. 
The NN outputs include the set of active constraints (i.e., constraints that are active in the optimal solution, this identifying saturated generators and transmission lines), as well as the non-zero load and wind curtailments. Third, using the NN outputs, the original DC-OPF problem is reduced to a linear system of equations, which can be solved efficiently. Fourth, based on the solution from the last step, the LMPs can be computed easily. The resulting LMPs are tested with respect to the requirements of efficient market clearing results to ensure their feasibility.
The training process is performed day-ahead so that the trained models can be used during the operating day in real time to solve DC-OPF based on a large number of possible configurations given updated forecasts and scenarios.
Table \ref{tbl:prior-works} compares the proposed PIMA-AS-OPF with three relevant ML approaches used to solve DC-OPF.
The proposed approach aligns with the theme of \textit{physics-informed} machine learning \cite{karniadakis2021physics} by directly taking advantage of the physics of the transmission grid and structure of the presently formulated real-time optimization procedure to construct an ML algorithm. Moreover, by testing the results against the crucial market properties such as revenue adequacy and cost recovery, the proposed method is also feasible for market implementations.

\begin{table*}[t]
\begin{center}
\caption{Comparison of Existing Approaches and the Proposed PIMA-AS-OPF Approach.}
\label{tbl:prior-works}
\begin{tabular}{p{2cm}p{4.8cm}p{4.8cm}p{4.8cm}}
\hline
\multicolumn{1}{c}{Article} & \multicolumn{1}{c}{Task}  &\multicolumn{1}{c}{Methodology}   & \multicolumn{1}{c}{Extension in PIMA-AS-OPF} \\
\hline
\multicolumn{1}{l}{Deka et al.~(2019) \cite{DeepMisra}} & Efficient solution of DC-OPF. & Learn the active set of constraints for the optimal DC-OPF solution.  & Substitute ML  with solving a system of linear equations when possible, and  solve dual problem. \vspace{4pt}\\
 \multicolumn{1}{l}{Liu et al.~(2021) \cite{liu2021graph}}  &  Topology-aware  solution of dual DC-OPF problem  with adherence to  constraints. &  Solve DC-OPF  using graph NN in order to predict LMPs. & Solve both  primal and  dual problem,  ensure LMPs  adhere to  financial coherency.\vspace{4pt}\\
 \multicolumn{1}{l}{Chen et al.~(2022) \cite{chen2022learning}} & Efficient learning of dual DC-OPF solution and primal objective. & Use NN to learn dual variables and LMPs. & Recover objective algebraically, check market conditions, validate on  large, realistic system.\vspace{1pt}\\
\hline
\end{tabular}
\end{center}
\end{table*}
\section{Preliminaries}
\label{sec:intro-tech}

This section describes the DC-OPF formulation, presents the Karush-Kuhn-Tucker (KKT) conditions of the DC-OPF formulation as used in the PIMA-AS-OPF algorithm, derives the expression for LMPs, and then clarifies the crucial properties of an efficient market.

\subsection{DC-OPF Formulation}\label{sec:DC-OPF}

Consider the following single-period DC-OPF model, which allows for load shedding and wind curtailment and computes power flows using the Power Transfer Distribution Factor (PTDF) matrix $\Phi$: 
\begin{subequations}
\begin{align}
\min_{p_g, s_i, c_w}
& \!\sum_g\!\Big(c_g^{(2)} p_g^2 + c_g^{(1)} p_g\Big) +\!\!\sum_i\! c^{\rm LNS}_i s_i+\!\! \sum_w\! c^{\rm RES}_w c_w, \label{eq:obj2}\\
\text{s.t.}\quad 
& \sum_i \left(\ell_i - s_i\right) = \sum_g p_g - \sum_w c_w  \label{eq:bal2}\,,\\
& \forall \, k \in L\colon \, -f_k^{\max} \leq \sum_i  \Phi_{ki} \Big[ \sum_{g \in G_i} p_g  - \sum_{w \in W_i} c_w  \nonumber \\
& \hspace{42pt}- \ell_i + s_i \Big] 
\leq f_k^{\max} \label{eq:flow2}\,,\\
& \forall \, g \in G\colon \, p_g^{\min} \leq p_g \leq p_g^{\max} \label{eq:gen2}\,,\\
& \forall \, i \in B\colon \, s_i \ge 0 \label{eq:load_slack}\,,\\
&\forall \, w \in W\colon \, c_w \ge 0 \label{eq:wind_slack}\,,
\end{align}%
\label{mod:DCOPF_PTDF}%
\end{subequations}%
where $B$, $L$, $G$, and $W$ are the sets of nodes, lines, conventional generators, and wind farms in the system, respectively, $G_i$ and $W_i$ are the sets of generators and wind farms connected to node $i$, $\ell_i$ is the forecast net load (i.e., forecast demand minus forecast renewable injections) at node $i$, $f_k^{\max}$ is the thermal capacity of line $k$, $p_g$ is the (active) power output of generator $g$, $p_g^{\min}$ and $p_g^{\max}$ are the upper and lower bounds of $p_g$, and 
$s_i$ and $c_w$ are slack variables representing the curtailed load at node $i$ and the curtailed wind power at wind farm $w$. Objective~\eqref{eq:obj2} minimizes the cost of power generation plus the penalties for load shedding and wind curtailment, where $c_g^{(2)}$ and $c_g^{(1)}$ are parameters in the quadratic cost function of generator $g$, and $c^{\rm LNS}_i$ and $c^{\rm RES}$ are the unit cost of curtailed load and wind, respectively. 
Note that the slack variables $s_i$ and $c_w$ should be capped by the forecast load and wind power, respectively, but those constraints are omitted for simplicity of the following manipulations since they are unlikely to be active. Adding these linear constraints will not affect the underlying method, as well as its computational complexity and applicability.

Solving the formulation~\eqref{mod:DCOPF_PTDF} effectively requires assuming that the unit commitment decisions on all conventional generators are known.

The proposed PIMA-AS-OPF algorithm can solve a number of real-time DC-OPF instances, each with their own configurations of wind power generation, therefore providing stochastic insights from a set of resulting solutions. 

The DC-OPF formulation in \eqref{mod:DCOPF_PTDF} can be re-stated as the following Lagrangian:
\begin{align}\label{eq:min-max}
& \min\limits_{\bm{p}}\max\limits_{\bm{d}} \mathcal{L}\big(\bm{p}, \bm{d}\big)\quad\text{s.t. Eq.~\eqref{eq:load_slack} and \eqref{eq:wind_slack}}, \bm{d} \setminus \lambda \geq 0,\\
&\mathcal{L}\big(\bm{p}, \bm{d}\big)  \coloneqq \sum_g \Big(c_g^{(2)} p_g^2 + c_g^{(1)} p_g\Big) + \sum_i  c^{\rm LNS}_i  \,s_i \nonumber \\ & +  \sum_w c^{\rm RES}_w  c_w+ \lambda \bigg[ \sum_i \big(\ell_i - s_i\big) - \sum_g p_g + \sum_w c_w\bigg] \nonumber \\
& + \sum_k\overline{\mu}_k \bigg[ \sum_i  \Phi_{ki} \bigg(\sum_{g \in G_i} p_g - \sum_{w \in W_i} c_w - \ell_i + s_i \bigg) - f_k^{\max}\bigg] \nonumber \\
& - \sum_k\underline{\mu}_k \bigg[ f_k^{\max} + \sum_i  \Phi_{ki} \bigg(\sum_{g \in G_i} p_g - \sum_{w \in W_i} c_w- \ell_i + s_i \bigg)\bigg] \nonumber \\
& + \sum_{g}\bigg[\overline{\alpha}_g \big( p_g - p_g^{\max} \big) + \underline{\alpha}_g \big( p_g^{\min} - p_g \big)\bigg]\,,
\label{eq:Lagrangian}
\end{align}
where $\bm{p} \coloneqq (p_g, s_i, c_w)$ denotes the vector of primal variables, and $\bm{d} \coloneqq (\lambda, \overline{\mu}_k, \underline{\mu}_k, \overline{\alpha}_g, \underline{\alpha}_g)$ denotes the vector of Lagrange multipliers associated with respective primal constraints.

\subsection{KKT Conditions of DC-OPF} \label{sec:KKT}

The PIMA-AS-OPF algorithm, described in Section \ref{sec:ML}, relies on the KKT conditions of the convex optimization in \eqref{eq:min-max}. The KKT conditions, which consist of \emph{complementary slackness} and \emph{stationarity at the optimum}, are necessary and sufficient conditions for the solution of a convex optimizaiton problem to be optimal \cite{beck2014introduction}. Complementary slackness states that a non-zero Lagrange multiplier exists only if the corresponding inequality constraint is active, i.e., satisfied with equality. Stationarity at the optimum refers to the partial derivative of the Lagrangian being zero when evaluated at the optimal solution. 
Accordingly, the KKT conditions of \eqref{eq:min-max} are given as follows:
\begin{subequations}
\begin{align}
& \textbf{Complementary Slackness:} \nonumber \\
& \forall k \colon \overline{\mu}_k \bigg[ \sum_i  \Phi_{ki} \bigg(\sum_{g \in G_i} p_g \!-\! \sum_{w \in W_i} c_w \!-\! \ell_i \!+ \!s_i \bigg) \!- \!f_k^{\max}\bigg] \!= \!0, \label{eq:mucs} \\
& \forall k \colon \underline{\mu}_k \bigg[ f_k^{\max}\! +\! \sum_i  \Phi_{ki} \bigg(\sum_{g \in G_i} p_g \!-\! \sum_{w \in W_i} c_w \!- \!\ell_i \!+ \!s_i \bigg)\bigg] \!= \!0, \label{eq:nucs} \\
& \forall g \colon \overline{\alpha}_g \big( p_g - p_g^{\max} \big)=0, \label{eq:gmaxcs} \\
& \forall g \colon \underline{\alpha}_g \big( p_g^{\min} - p_g \big)=0, \label{eq:gmincs} \\
& \textbf{Stationarity:} \nonumber \\
& \forall g \colon  2c_g^{(2)} p_g \!+\! c_g^{(1)} \!-\! \lambda \!+\! \sum_k \!\Phi_{ki} \big(\overline{\mu}_k \!-\! \underline{\mu}_k\big)\!+\!\overline{\alpha}_g \!\!-\!\! \underline{\alpha}_g\!=\! 0, \label{eq:genst} \\
& \forall i \colon\, c^{\rm LNS}_i \!-\! \lambda \!+\! \sum_k\! \Phi_{ki} \big(\overline{\mu}_k \!-\! \underline{\mu}_k\big) \!=\! 0 \label{eq:lsst} \,,\\
& \forall w \colon \, c^{\rm RES}_w \!+\! \lambda \!+\! \sum_k\! \Phi_{ki} \big(\underline{\mu}_k \!-\! \overline{\mu}_k\big) \!=\! 0,\label{eq:wst}
\end{align}%
\label{mod:KKT}%
\end{subequations}%

\subsection{LMPs and Market Design Properties}
\label{sec:LMP_properties}
LMPs in electricity markets reflect the cost of producing, as well as the price of consuming, an additional MWh of electricity at a given bus. The LMP at node $i$ can be derived by differentiating the Lagrangian in Eq.~\eqref{eq:Lagrangian} with respect to net load $\ell_i$, as:
\begin{equation}
\forall i:\quad \mathrm{LMP}_i \coloneqq \frac{\partial \mathcal{L}}{\partial \ell_i} = \lambda + \sum_k \Phi_{ki} \big(\underline{\mu}_k - \overline{\mu}_k\big).
\label{eq:LMP}
\end{equation}
Specifically, $\lambda$ is the LMP at the reference node, and it is also the LMP at all nodes when the transmission capacity is unlimited, i.e., when $\underline{\mu}_k$ and $\overline{\mu}_k$ are zero. 

LMPs should have the following critical properties to ensure sound electricity market operations \cite{Wolak2000}:
\subsubsection{Revenue Adequacy}
From the perspective of the system operator, the total payment made by the market to the producers must be covered by the total payment made to the market by consumers.
In addition, recall the system operator must pay the designated penalties for load shedding and wind curtailment. While not directly enforced during real-time operations, these penalties represent the unanticipated cost to the system operator, e.g., of having to rely more heavily on traditional generation technologies when wind power is curtailed. Thus, the  revenue adequacy condition is formulated as:
\begin{align}
&\sum_g \text{LMP}_{i(g)} p_g + \sum_i c^{\rm LNS}_i s_i + \sum_w c^{\rm RES}_w c_w\nonumber\\ &\leq \sum_i \text{LMP}_i \big(\ell_i - s_i\big)\,.
\label{eq:rev-adeq-2}
\end{align}

\subsubsection{Cost Recovery}

From the perspective of individual generators, market participants must recover their operating costs based on the resulting LMPs, i.e., the payment made to each generator must be sufficient to recover their operating costs. Therefore, the cost recovery condition is formulated as follows:
\begin{equation}
\forall \, g \colon \, 2c_g^{(2)} p_g^2 + c_g^{(1)} p_g \leq \text{LMP}_{i(g)} \, p_g\,.
\label{eq:cost-rec}
\end{equation}
For generators which do not reach their upper or lower limits (so-called marginal generators), Eq.~\eqref{eq:cost-rec} holds with equality. 

\subsubsection{Efficiency via Strong Duality}
Since DC-OPF as stated in Eq.~\eqref{mod:DCOPF_PTDF} is a quadratic, convex optimization problem with all affine constraints, the strong duality is guaranteed, i.e.,  the primal optimal objective and the dual optimal objective are equal. 
In the unlikely event that $s_i = \ell_i$ or $c_w = \omega_i$, in principle, an additional nonzero Lagrangian multiplier is activated and must be accounted for in the Lagrangian and strong duality condition.

\section{PIMA-AS-OPF Algorithm}
\label{sec:ML}
In this section, an ML approach (PIMA-AS-OPF) is proposed to enhance the computational efficiency of solving the DC-OPF given in~\eqref{mod:DCOPF_PTDF}.
The proposed PIMA-AS-OPF procedure includes four steps: 
\begin{enumerate}
\item Generate multiple DC-OPF instances using a standard DC-OPF optimization solver and use these as ground truth for the subsequent machine learning steps.
\item Train a Neural Network (NN) to identify the active constraints and the non-zero slack variables in DC-OPF. The loss function for NN training is designed in Section~\ref{sec:loss}, while the NN inputs and outputs are presented in Section~\ref{sec:input+output}.
\item Build an equivalent system of equations based on the NN outputs, which provides the same optimal solution as the DC-OPF problem, but is more computationally amenable. This equivalent system of equations is derived in Section \ref{sec:lin-eqs}. 
\item Use the derived relationship in \eqref{eq:LMP} to compute the resulting LMPs. Test the resulting LMPs against the market properties in Section~\ref{sec:LMP_properties} such as revenue adequacy and cost recovery to validate their feasibility. This result validation procedure is discussed in Section~\ref{sec:market}.
\end{enumerate}

This approach exploits the fact that the number of active constraints in DC-OPF is relatively small, which is supported by NYISO's findings in \cite{NYISO} that only about 15 out of the 2203 lines in the system are regularly operated at their nominal thermal limits. 

\subsection{NN Inputs and Outputs}\label{sec:input+output}
Active set learning can be interpreted as a supervised classification problem where an ML model, e.g., a NN, is trained to assign a binary label ($0/1$) to each relevant variable. The physical meaning of labels varies based on each specific application. 

The NN inputs consist of the information accessible to ISOs during the real-time scheduling process. This information includes the unit commitment results and the forecast demand and renewable energy generation, which can be used to calculate the net load ($\ell_i$). This information, however, requires pre-processing before being provided to the NN. Specifically, the ``load flows" are  computed as explained below based on the given net load and used as an additional input to the NN. Recall that the power flow on line $k$ can be expressed as:
\begin{equation}
f_k = \sum_{i \in k_i} \Phi_{ki} \bigg[\sum_{g \in G_i} p_g - \sum_{w \in W_i} c_w - \ell_i + s_i \bigg].
\label{eq:flow}
\end{equation}
Although the values of $p_g$, $s_i$, and $c_w$ in \eqref{eq:flow} are unknown before the DC-OPF is solved, a proxy for the power flows in the system can be computed based on $\ell_i$ and the topology of the system as encoded in the PTDF matrix $\Phi$. Therefore, the \emph{load flow} on line $k$ is defined as:
\begin{equation}
f_k^{\rm load} \coloneqq \sum_{i \in k_i} \Phi_{ki} \ell_i\,.
\label{eq:loadflow}
\end{equation}
Both the load at each node and the load flow on each line are provided as inputs to the NN. In Section \ref{sec:case-study}, it is demonstrated that the addition of load flows leads to improved replication of dispatch decisions and LMPs.

Classic active set learning methods for solving DC-OPF can identify the saturated lines and generators, see e.g., \cite{DeepMisra,misra2022learning}. However, our approach goes further by also determining if pre-arranged adjustments, such as load shedding or wind curtailment, are necessary to make the DC-OPF problem feasible. As a result, the desired NN outputs  include:
\begin{enumerate}
\item The committed generators that reach their upper or lower output limits, 
\item The transmission lines that reach their thermal limits,
\item The buses with non-zero load shedding, and 
\item The wind farms with non-zero wind power curtailment.
\end{enumerate}

Given the commitment status of generators, all desired outputs can be determined given the net load and the topology of the system, which ensures the feasibility of the NN model.
The NN output is a matrix that consists of binary values. Since a Lagrange multiplier is zero if and only if the corresponding constraint is inactive, the NN outputs will include the Lagrange multipliers associated with constraints of generator outputs and transmission capacities, i.e., Eqs.~\eqref{eq:gen2} and \eqref{eq:flow2}. Meanwhile, the NN outputs should also include binary variables indicating whether load or wind power is curtailed, where ``1'' denotes ``with curtailment'' and ``0'' denotes ``without curtailment.''

\subsection{Loss Function for NN Training}\label{sec:loss}

In the PIMA-AS-OPF algorithm, the NN is used to identify the active constraints of generators and transmission lines in the DC-OPF formulation, as well as the occurrences of load shedding and wind curtailment. Accordingly, the label of ``1'' means a constraint is active or a slack variable is non-zero, and vice versa. 
Following conventional classification models such as logistic regression, the following loss function is designed for NN training:
\begin{equation}
L \coloneqq \sum_{s,v} \big| \max_s d_v^s \big|\, L_{\rm reg} \Big(\bm{y}_v^{s}, \mathcal{B}\big(d_v^s\big)\Big)\,.
\label{eq:loss}
\end{equation}
In Eq.~\eqref{eq:loss}, 
vector $d_v$ denotes the value of variable $v$, which contains all of the variables of interest, such as the Lagrange multipliers of relevant constraints in DC-OPF). The values of $d_v$ are examined at each sample of the testing set $s$. The function $\mathcal{B}(d)$ is a binary mapping derived from $d$. For example, if $\overline{\mu}_k \neq 0$, then $\mathcal{B}(\overline{\mu}_k) = 1$; else $\mathcal{B}(\overline{\mu}_k) = 0$. Meanwhile, matrix $\bm{y}$ denotes the neural network outputs, which are between $0$ and $1$, and are compared to the ground truth values given by $\mathcal{B}(d_v^s)$. The components of $\bm{y}$ are assessed using a pre-set threshold, e.g., 0.5, to determine whether or not the corresponding constraints are active. The function $L_{\rm reg}(x,y)$ measures the difference between $x$ and $y$, as given by:
\[
L_{\rm reg} \big(x,y\big) = \begin{cases}
-\log\big(x\big), & y = 1\,; \\
-\log\big(1-x\big), & y = 0\,.
\end{cases}
\]
Therefore, $L_{\rm reg} \left(\bm{y}_v^{d(s)}, \mathcal{B}(d_v^s)\right)$ represents the difference between the NN outputs  and its true value with respect to the $s^\text{th}$ sample in the training set and the $v^\text{th}$ variable of interest. In addition, this difference is weighted by the maximum value of $d_v$ across the training data set. These weights place greater emphasis on penalizing the NN for incorrect identification of active constraints with large Lagrange multipliers, as this can cause significant errors in LMPs whose values are computed based on the Lagrange multipliers (see Eq.~\eqref{eq:LMP}). Meanwhile, more weight will be placed on variables that are not consistently zero or non-zero, as they display less discernible patterns are therefore more challenging for the NN to learn. This weighting method is specifically designed for the task of identifying active constraints since it takes into account the underlying structure of the DC-OPF formulation.

\subsection{Equivalent System of Equations}\label{sec:lin-eqs}
In what follows, the ``free'' generators are defined as the generators with none of their output limits  saturated. Similarly, the ``free'' lines are the transmission lines with none of their thermal limits are saturated.

Based on the NN outputs, all the zero elements in the vectors of Lagrange multipliers $\{\overline{\alpha}_g, \underline{\alpha}_g\}_{g \in G}$ and $\{\overline{\mu}_k,\underline{\mu}_k\}_{k \in L}$, as well as the zero elements in the vectors of slack variables $\{s_i\}_{i \in B}$ and $\{c_w\}_{w \in W}$, can be obtained. Therefore, the remaining unknown variables include the \emph{non-zero} primal variables ${p_g}$, $s_i^{*}$, $c_w^{*}$ and the \emph{non-zero} Lagrange multipliers $\overline{\mu}_k^*$, $\underline{\mu}_k^*$, $\overline{\alpha}_g^*$ and $\underline{\alpha}_g^*$, where superscript $^*$ denotes non-zero values. With this information, the original DC-OPF formulation can be reduced to the following equivalent system of equations (ESE):
\begin{subequations}
\begin{align}
& \forall \overline{\mu}_k^*\colon \!\sum_i\!  \Phi_{ki} \bigg[\sum_{g \in G_i}\! p_g \!-\! \sum_{w \in W_i}\! c_w \!-\! \ell_i \!+\! s_i \!\bigg] \!=\! f_k^{\max}, 
\label{eq:equ_mucs} 
\\
& \forall \underline{\mu}_k^*\colon \!\sum_i\!  \Phi_{ki} \bigg[\sum_{g \in G_i} \!p_g \!-\! \sum_{w \in W_i}\! c_w\!-\! \ell_i \!+\! s_i \bigg] \!=\! -f_k^{\max}, \label{eq:equ_nucs} 
\\
& \forall \overline{\alpha}_g^*\colon \!p_g = p_g^{\mathrm{max}}, \label{eq:equ_gmaxcs} 
\\
& \forall \underline{\alpha}_g^*\colon \,p_g = p_g^{\mathrm{min}}, \label{eq:equ_gmincs} 
\\
& \forall g \in G_{\rm free} \colon  2c_g^{(2)} p_g \!+\! c_g^{(1)} \!-\! \lambda \!+\! \sum_k \!\Phi_{ki} \big(\overline{\mu}_k \!-\! \underline{\mu}_k\big)\!=\! 0, 
\label{eq:equ_genst} 
\\
& \forall s_i\colon\, c^{\rm LNS}_i \!-\! \lambda \!+\! \sum_k\! \Phi_{ki} \big(\overline{\mu}_k \!-\! \underline{\mu}_k\big) \!=\! 0 
\label{eq:equ_lsst} \,,
\\
& \forall c_w\colon \, c^{\rm RES}_w \!+\! \lambda \!+\! \sum_k\! \Phi_{ki} \big(\underline{\mu}_k \!-\! \overline{\mu}_k\big) \!=\! 0  \label{eq:equ_wst}, \\
& \sum_i \left(\ell_i - s_i\right) = \sum_g p_g - \sum_w c_w, \label{eq:equ_bal2}
\end{align}%
\label{mod:equ_system}%
\end{subequations}%
where Eqs.~\eqref{eq:equ_mucs}-\eqref{eq:equ_gmincs} are the complementary slackness conditions of the identified active constraints, Eq.~\eqref{eq:equ_genst} is the stationarity of free generators, Eqs.~\eqref{eq:equ_lsst} and \eqref{eq:equ_wst} are the same as the the stationarity in Eqs.~\eqref{eq:lsst} and \eqref{eq:wst}, and Eq.~\eqref{eq:equ_bal2} is the same as the power balance constraint in Eq.~\eqref{eq:bal2}. Since the objective of the PIMA-AS-OPF algorithm is to calculate the LMPs, the non-zero Lagrange multipliers that are not used in the expression of LMPs in Eq.~\eqref{eq:LMP}, such as $\overline{\alpha}_g^*$ and $\underline{\alpha}_g^*$, can also be eliminated from the ESE to reduce the dimensionality of the problem. In summary, the dimensionality of the ESE is $N \times N$, where $N = |k \, \colon \, \overline{\mu}_k \neq 0| + |k \, \colon \, \underline{\mu}_k \neq 0| + |G| + |i \, \colon \, s_i > 0| + |w \, \colon \, c_w > 0| + 1$. This is a significant reduction from the dimensionality of the original DC-OPF problem, which is $3|G| + 2|L| + |B| + |W| + 1$, with $|G|, |L|, |B|, |W|$ denoting the total number of generators, lines, buses, and wind farms, respectively.

\subsection{LMP Validation Based on Market Properties}
\label{sec:market}

For a ML approach to be adopted by real-world system operators, it is important to ensure the feasibility of its output. As discussed in Section~\ref{sec:LMP_properties}, efficient market clearing results should have the properties of revenue adequacy, cost recovery, and efficiency. Therefore, to validate the effectiveness of the resulting LMPs, they need to be tested against selected market properties, such as Eqs.~\eqref{eq:rev-adeq-2} and \eqref{eq:cost-rec}. In the proposed PIMA-AS-OPF algorithm, these conditions follow as direct consequences of the DC-OPF formulation. However, the learning  algorithm is designed in such a way as to be robust to ``small errors''. Thus, it is anticipated that, if a saturated line, generator, or another constraint is misclassified, and the associated Lagrange multiplier is small in magnitude, the market design properties are still preserved. This claim is verified in the case study presented in Sec.~\ref{sec:case-study}.

\section{Case Study in NYISO 1814-Bus System}\label{sec:case-study}

\begin{figure}[h]
\center
\includegraphics[width=0.9\columnwidth]{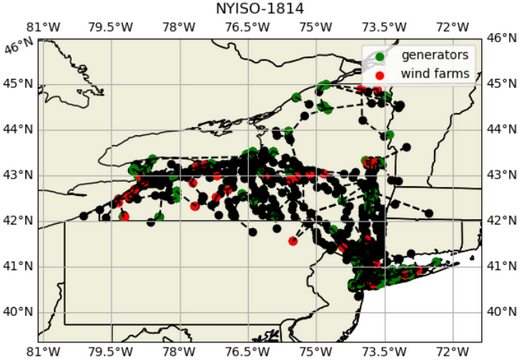}
\caption{The NYISO-1814 system.}
\label{fig:NYISOmap}
\end{figure}

The effectiveness of the proposed PIMA-AS-OPF algorithm  is demonstrated on the 1814-bus NYISO system, which is referred to as NYISO-1814 in this section. The NYISO-1814 system, as shown in Fig.~\ref{fig:NYISOmap}, 
consists of 1814 buses, 2203 lines, 362 conventional generators, and 33 wind farms, and it is sufficiently large to test the PIMA-AS-OPF scalability.

\subsection{Simulation Setup}
\label{sec:setup}
The ground truth DC-OPF solutions are generated using Gurobi 10.0 \cite{gurobi} via JuMP \cite{jump} in the Julia programming language. The neural networks for classification are constructed and trained using the Flux.jl package \cite{Flux1,Flux2}. All other computations are performed using the native functions of Julia, or with the LinearAlgebra.jl package.

All the NN models in this case study share the same architecture, consisting of an input layer, three hidden layers with 30 neurons each and a ReLU activation function, and an output layer with a sigmoid activation function. The training of the NN models is performed on a high-performance computer, and it takes approximately 15-20 min to train each NN with a maximum training duration of 1000 epochs. In practice, the real-time system operators will be provided with a ready-to-use NN trained in advance based on the unit commitment results of day-ahead scheduling, and thus will only be required to perform the ``system of linear equations'' phase (as in Fig. 1) of the PIMA-AS-OPF approach in real time. 

The testing of the NN is performed on a Lenovo IdeaPad with 12 GB RAM and an Intel Core i5 10 Gen processor. In summary, the testing results show that DC-OPF is solved by Gurobi in 0.11-0.13 sec, while the equivalent system of equations is solved in only 0.04-0.05 sec.

\subsection{Six Basic Cases}
\label{sec:UC}
To assess the effectiveness of the scheme, DC-OPF is solved for three representative combinations of load and wind injections obtained from two days in February and August, respectively, at different times.
Furthermore, for each time, a \textit{base} and \textit{high} wind profile is considered to capture both current and projected wind power outputs under increased renewable penetration.  Accordingly, Table~\ref{tab:cases} summarizes the six unit commitment (UCs)  instances used to solve Eq.~(\ref{eq:min-max}) for the NYISO system. Each unit commitment instance is solved \textit{a priori}, resulting in a schedule of generators to be  \textit{on} and \textit{off} when DC-OPF is solved in real-time. The three selected times constitute conditions under which the number of committed generators is small, moderate, and large. Cases A7B and A7H represent typical early-morning conditions, with 273 out of 362 generators committed. By contrast, the evening commute brings greater demand and necessitates greater generation capacity throughout the NYISO system, leading to 313 out of 362 commitments for A17B and A17H. In addition, an unanticipated thunderstorm occurred in the vicinity of Long Island at this time, presenting additional challenges with respect to uncertainty from wind farms. Finally, demand is relatively low in the early morning, so relatively few generators are online for F0B, with even fewer required with a higher wind power output level in F0H. 

\begin{table}[!t]
    \caption{Overview of Six Basic Cases.}
    \begin{center}
    \begin{threeparttable}
        \begin{tabular}{ccccccc}
            \toprule
                Case & Time & Wind & On\tnote{1} & $P_{\min}$\tnote{2} & $P_{\max}$\tnote{3}& $P_{\rm sched}$\tnote{4}\\ 
            \midrule
                A7B & Aug.28, 7am & Base & 273 & 87.49 & 289.41 & 231.10\\
                A7H & Aug.28, 7am & High & 273 & 87.49 & 289.41 & 231.10\\
                A17B & Aug.28, 5pm & Base & 313 & 102.53 & 336.59 & 312.79\\
                A17H & Aug.28, 5pm & High & 313 & 102.53 & 336.59& 312.79\\
                F0B & Feb.13, 12am & Base & 187 & 55.78 & 184.71 & 142.67\\
                F0H & Feb.13, 12am & High & 141 & 50.81 & 158.43 & 123.89\\
            \bottomrule
        \end{tabular}%
        \begin{tablenotes}
            \item[1] Number of committed generators in this case
            \item[2] Minimum output of all committed generators (unit: MW)
            \item[3] Maximum output of all committed generators (unit: MW)
            \item[4] Scheduled output of all committed generators  (unit: MW)
        \end{tablenotes}
    \end{threeparttable}
    \end{center}
    \label{tab:cases}
\end{table}

\subsection{Sample Generation Procedure}
\label{sec:sampling}
Training input data samples are constructed from the given wind injections by adding tuned Gaussian noise.
For each wind farm $i$, wind power injection $w_i^0$ is perturbed as:
\begin{equation}
w_i = w_i^0(1 + \eta\, \xi_i), \quad \xi_i \sim\mathcal{N}(0,1),
\label{eq:wind-perturb}
\end{equation}
Noise level $\eta$ is set to 0.01, 0.05, 0.1, and 0.15, increasing from a negligible level to one consistent with typical wind power variance \cite{LIPMAN1982149,DvorkinWind}. Note that load is not varied for this particular phase of the case study, as most of the variability arises from wind. The combination of changing unit commitment, wind penetration, and noise configurations introduced to the wind farms on top of the pre-set conditions presents varying levels of challenge to the neural network with respect to learning saturated generators, saturated lines, shed loads, and curtailed wind injections. Hence, a broad range of operating decisions can be captured. ML tools will be exposed to these decisions in real-world operations.

For each case in Table~\ref{tab:cases}, 1000 samples are generated, half of which are allocated for training and the rest for testing. A separate neural network is trained for each configuration. However, all neural networks share the same architecture from Section~\ref{sec:setup}.

\begin{remark} The complementary slackness conditions for the generator constraints are omitted, as saturated generators are fixed during the DC-OPF solution. As such, stationarity conditions are only written for free generators $g \in G_{\rm free}$. Generators which reach their upper or lower limits in real-time are learned by the NN. Since these limits are known, the corresponding $p_g$ values are assumed fixed for the purpose of constructing the system of linear equations, and are subtracted from the net load $\ell_i$ at the bus where they are located prior to solving the system. 
\end{remark}

\subsection{Accuracy of Classification}
\label{sec:classification}

The success of the proposed ML-driven OPF scheme depends on the correct classification of saturated generators, saturated lines, shed loads, and curtailed wind farms by the neural network. For example, misidentifying a saturated line with a high-magnitude Lagrange multiplier as below its limit by the neural network will result in large LMP errors at the attached and possibly nearby buses. On the other hand, by analyzing the results of the neural network classification, LMP and dispatch errors may be localized to particular buses, lines, generators, and wind farms, informing strategic improvements to the machine learning scheme to target the most vulnerable components of the power grid.
\begin{table}[H]
\centering
\caption{Percentage of misclassifications in different test cases.}
\begin{threeparttable}
\begin{tabular}{lrrrrr} 
 \toprule
\multicolumn{1}{c}{Case} & \multicolumn{1}{c}{Noise $\eta$} & \multicolumn{1}{c}{Generators\tnote{1}} & \multicolumn{1}{c}{Lines\tnote{2}} & \multicolumn{1}{c}{Load\tnote{3}} & \multicolumn{1}{c}{Wind\tnote{4}}\\ 
 \midrule
A7B & 0.01 & 1.10\% & 0  & 0 & 0 \\
 & 0.05 & 1.00\% & 0  & 0 & 0 \\
 & 0.1 & 0.88\% & 0  & 0 & 0 \\
 & 0.15 & 0.82\% & 0  & 0 & 0 \\
 \midrule
A7H & 0.01 & 0  & 0.01\% & 0 & 0 \\
 & 0.05 & 0.42\% & 0.02\% & 0 & 0\\
 & 0.1 & 1.14\% & 0.02\% & 0 & 0.04\%\\
& 0.15 & 1.90\% & 0.03\% & 0 & 0.29\%\\
 \midrule
 A17B & 0.01 & 0.05\% & 0.001\% & 0 & 0\\
 & 0.05 & 0.42\% & 0.02\% & 0 & 0 \\
 & 0.1 & 0.72\% & 0.02\% & 0 & 0 \\
& 0.15 & 0.92\% & 0.02\%  & 0 & 0 \\
 \midrule
  A17H & 0.01 & 0 & 0 & 0 & 0 \\
 & 0.05 & 0.10\% & 0.0005\% & 0 & 0.03\% \\
 & 0.1 & 0.47\% & 0.02\%  & 0 & 0.57\%\\
& 0.15 & 1.21\% & 0.01\% & 0 & 1.18\%\\
 \midrule
   F0B & 0.01 & 0.04\% & 0.02\% & 0 & 1.18\% \\
 & 0.05 & 0.41\% & 0.02\% & 0 &  1.41\%\\
 & 0.1 & 0.89\% & 0.02\% & 0 & 1.45\% \\
& 0.15 & 1.49\% & 0.03\% & 0 &  1.45\% \\
 \midrule
    F0H & 0.01 & 1.43\% & 0.002\% & 0 & 0.08\% \\
 & 0.05 & 5.79\% & 0.02\% & 0 & 1.42\% \\
 & 0.1 & 8.45\% & 0.03\% & 0 & 2.47\% \\
& 0.15 & 9.43\% & 0.04\% & 0 & 3.42\%\\
 \bottomrule
 \end{tabular}%
        \begin{tablenotes}
            \item[1] Misclassification of committed generators being free or non-free 
            \item[2] Misclassification of transmission lines being free or non-free
            \item[3] Misclassification of load shedding being zero or non-zero
            \item[4] Misclassification of wind curtailment being zero or non-zero
        \end{tablenotes}
    \end{threeparttable}
    \label{tbl:misclass}
\end{table}
Table \ref{tbl:misclass} displays the percentage of each quantity of interest which are misclassified by the neural network itemized for each configuration and noise level. The following observations are highlighted:
\begin{enumerate}
\item Across all configurations, as the noise associated with the wind perturbation increases, the number of misclassified lines and wind farms increases. This is the result of more volatile line congestion patterns and of increased wind power at farm locations with curtailment.
\item The neural network is more successful for the base wind than for the high wind configurations, 
which results  in more frequent curtailment in the latter case. 
\item Overall performance of the neural network is notably worse for F0B and F0H than for the other configurations. 

These two configurations correspond to a time of 12:00 midnight, when wind is typically higher.  This results in a surplus production of wind power which, especially when combined with the higher levels of volatility, leads to less predictable patterns in wind curtailment that are missed by the neural network. Indeed,  misclassifications are evident by the presence of infeasible wind curtailment values in the F0H case. Furthermore, the effect of adding $\eta$ level of noise to the wind profile in F0B and F0H results in a higher level of relative noise to the net load, which is provided as an input to the neural network and thus affects the accuracy of classifications. Combining the high wind profile with the highest level of noise, performance of the neural network, and consequently the system of equations, is unsatisfactory for these configurations.
\item Misclassification of saturated generators is less consistent than for the other quantities. This is not necessarily damaging to the ultimate OPF solution, as the system of equations may still compute a dispatch near the generator capacity if e.g., a saturated generator is considered free. Similarly, classifying a generator near, but not exactly equal to its limit as saturated is not directly harmful to the quality of the LMP and dispatch replication. For instance, in testing trial 98 in the A17H test case, all of the $p_g$ discrepancies are below 0.25 MW p.u, and at least one correctly classified generator exhibits a larger error than the misclassified generators do.
\item Load shedding is identified with complete accuracy across the board for all configurations. 
The load shedding penalty $c_i^{\mathrm{LNS}}$ is chosen such that it is unique for every bus to avoid ambiguous DC-OPF solutions, which results in more consistent load shedding patterns to be learned by the neural network.
\end{enumerate}

\begin{figure*}[t]
\centering
\includegraphics[width= \textwidth]{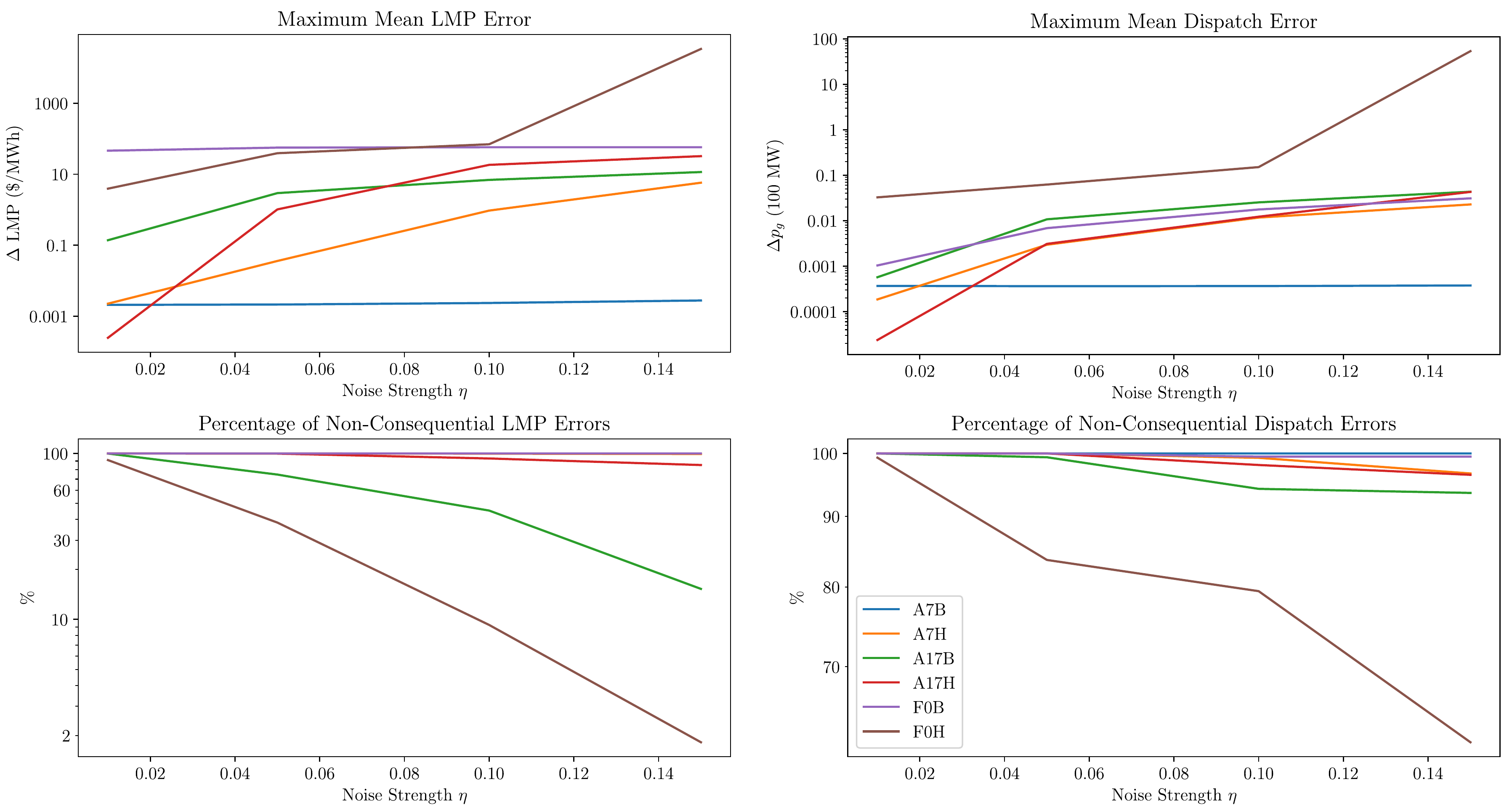}
\caption{Errors in reproduced LMPs ($\Delta \text{LMP}$) and dispatch of free generators ($\Delta p_g$). The errors are defined as consequential when $|\Delta \text{LMP}|>0.1$ \$/MWh and $|\Delta p_g|>0.01$p.u., respectively. }
\label{fig:err-graphs}
\end{figure*}

\subsection{Accuracy of DC-OPF Solution}\label{sec:accuracy}

\begin{figure*}[t]
\includegraphics[width= \textwidth]{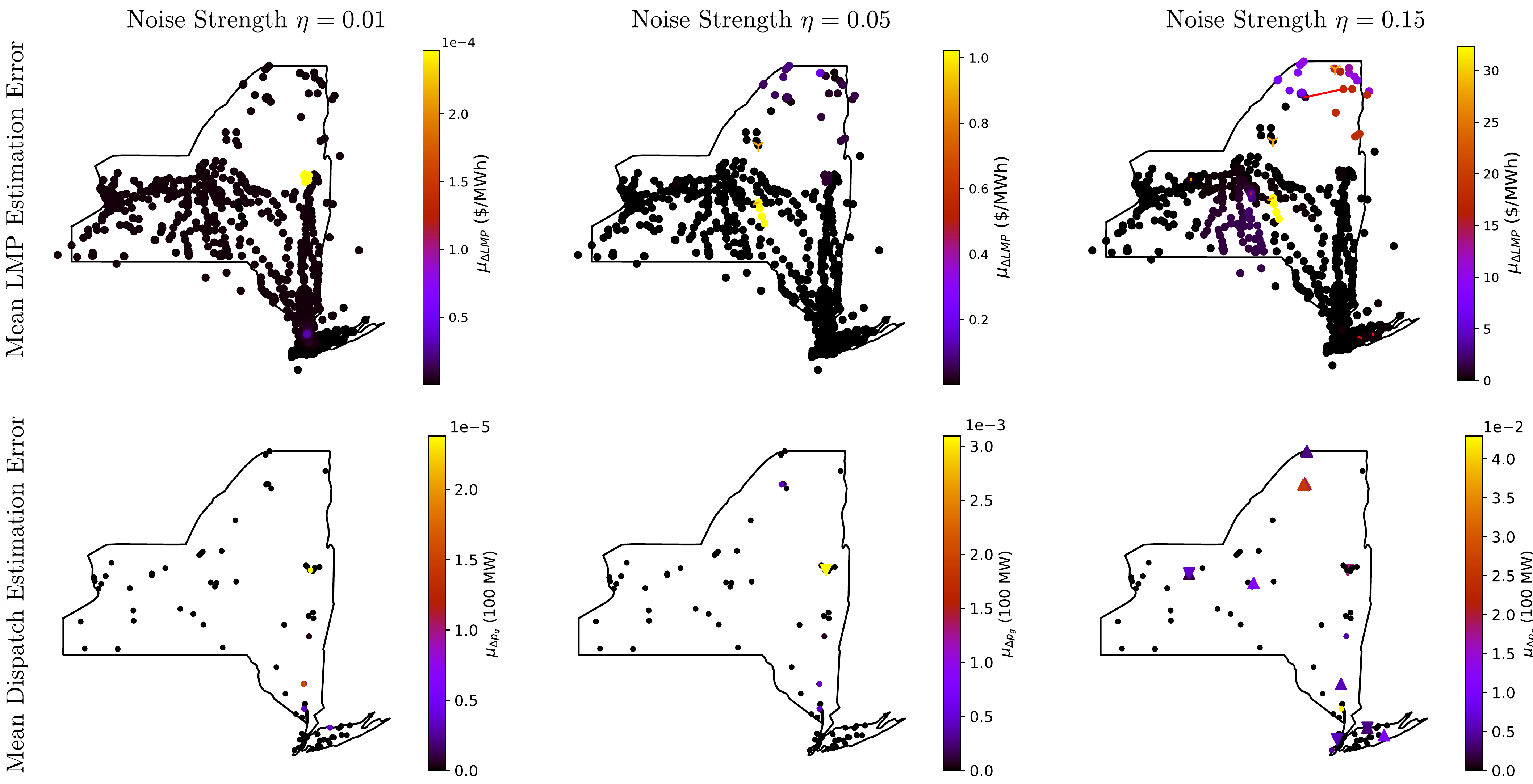}
\caption{Mean LMP and $p_g$ errors for the A17H configuration.  Misclassified lines and wind farms indicated are display, in top panels, in red for false negatives are red and in orange for false positives. Consequential misclassified generators (i.e. when dispatch error is greater than $0.001$ p.u.) are displayed, in bottom panels, as $\blacktriangledown$ for false negatives and as $\blacktriangle$ for false positives.}
\label{fig:err-maps-pg}
\end{figure*}

Both the primal (dispatch of generators) and dual (LMPs) solution of DC-OPF are of interest. 
 Figure~\ref{fig:err-graphs} depicts the error in learned LMPs and dispatch using the mean absolute error across the testing set. Subfigures \ref{fig:err-graphs}(a) and \ref{fig:err-graphs}(b) show the maximum mean error, i.e., the error at the bus or (free) generator experiencing the largest discrepancy, while subfigures \ref{fig:err-graphs}(c) and \ref{fig:err-graphs}(d) show the percentage of buses or generators with mean errors below a specified threshold. These measures of success are consistent with those for the neural network. Namely, as neural network performance improves, so does the accuracy of computation of dispatch and LMPs. Indeed, for most of the inspected configurations, over 90\% of buses and generators exhibit mean errors of an acceptable magnitude, while maximum errors may be identified and explained in an interpretable way. Notably, the F0H configuration, at high levels of noise, exhibits unsatisfactory results by all measures, with some LMP errors upwards of \$30,000/MWh, which 
coincides with misclassification of curtailment in this case.
As previously mentioned, it is possible to interpret the buses and generators experiencing large errors. Figure~\ref{fig:err-maps-pg} provides heat maps for the A17H configuration at all levels of noise, with frequently misclassified lines, wind farms, and generators emerging at high levels of noise and indicated on the maps. The magnitudes of errors increase as noise increases. In addition, the largest LMP errors are concentrated in ``clusters'' of buses, and each cluster may be associated with a misclassified line or wind farm. Since the entire set of misclassified entities is small, NNs can be adapted by introducing penalty terms to the loss function to target misclassification of those particular lines and wind farms. This aligns with the notion of ``physics-informedness,'' which, in physics-informed neural networks, typically entails adding terms to the loss function to enforce constraints derived from the underlying physics of the system (e.g. satisfaction of an ODE) \cite{karniadakis2021physics}.

\subsection{LMP Validation Results}\label{sec:financial}

To conclude the case study, the fidelity of the testing set with respect to selected market design conditions is assessed. The goal of PIMA-AS-OPF is to replicate the ground-truth DC-OPF in all respects as accurately as possible. As the ground-truth DC-OPF is convex, efficiency via strong duality holds by construction, as long as the neural network classification step is successful. Thus, empirical results are provided for revenue adequacy and cost recovery. Furthermore, cost recovery is only assessed for the ``free'' generators in each sample. 
Considering all sample configurations except for F0H, revenue adequacy is satisfied in all samples. This demonstrates that, under unit commitments for which DC-OPF solution via PIMA-AS-OPF is a reasonable approach, even when the neural network does not perform perfectly, the desired market design conditions are still satisfied. In the F0H configuration, which exhibited the least satisfactory results by far,  
all samples satisfy revenue adequacy by \eqref{eq:rev-adeq-2}. In addition, considering all free generators across the testing set, 195 out of 23767 (0.82\%) do not satisfy cost recovery with equality. Revenue adequacy in both forms is satisfied for all samples in F0B, with only 3 out of 240000 (0.0125\%) of free generators failing to satisfy cost recovery. 

\section{Conclusion and Future Work}
This paper bridges efforts in the power systems community to accelerate optimization using machine learning with the physical constraints and economic properties of electricity markets. The case study on NYISO-1814 demonstrates that the NN in the PIMA-AS-OPF pipeline can accurately identify the saturated lines, saturated generators, shed loads, and curtailed wind farms, for different sets of unit commitments and various levels of wind volatility. The outputs of the NN can then be used to reproduce the dispatch and LMPs using a system of linear equations, and the resulting LMPs adhere to the principle of revenue adequacy and cost recovery. However, some particular sample configurations may not be replicated effectively. Nonetheless, for these cases, specific features of the NYISO system that contribute to these deficiencies can be identified.

The optimization procedure in the DC-OPF problem is not completely replaced by ML in this study. Instead, ML is used to leverage specific properties of the optimization constraints, such as discrete degrees of freedom (such as saturated lines and generators, loads shed, and renewable generators curtailed) and sparsity, while preserving the physics underlying power system operations. In other words, the ML algorithm is only used to learn the infrequent discrete decisions. The remaining DC-OPF solution is computed using a system of linear equations, which is an exact method and more computationally efficient than optimization.

Future work will involve modifying the NN architecture and adjusting hyper-parameters, particularly when the current NN is not performing satisfactorily. Additionally, a similar ML approach for the unit commitment problem will be developed, which poses unique challenges due to its mixed-integer programming nature. The goal is to create an integrated learning-based pipeline for the joint UC-OPF problem that can be used for both day-ahead and real-time market operations.

\section*{Acknowledgment}

The authors would like to collectively acknowledge with gratitude the ARPA-E PERFORM grant from which this work arose. The UArizona team acknowledges complementary funding from the office of the Vice President for Research. The work of RF and MC on the project was partially supported by the UArizona NSF RTG in Data-Driven Discovery. 


\end{document}